\documentclass{article}
\usepackage{ijcai25}

\usepackage{times}
\usepackage{soul}
\usepackage{url}
\usepackage[hidelinks]{hyperref}
\usepackage[utf8]{inputenc}
\usepackage[font=footnotesize,skip=5pt]{caption}
\usepackage{graphicx}
\usepackage{amsmath}
\usepackage{amsthm}
\usepackage{booktabs}
\usepackage{algorithm}
\usepackage[switch]{lineno}

\usepackage{latexsym}

\newtheorem{remark}{Remark}
\newtheorem{definition}{Definition}
\newtheorem{example}{Example}
\newtheorem{corollary}{Corollary}
\newtheorem{theorem}{Theorem}

\usepackage{xcolor}
\usepackage{amsmath}
\usepackage{xspace}
\usepackage{amssymb}

\DeclareMathOperator*{\argmax}{arg\,max}

\definecolor{nicegreen}{rgb}{0.1,0.5,0.1}

\newcommand{\prob}[1]{\ensuremath{\mathbb{P}({#1})}}

\newcommand{\States}{\ensuremath{{\cal S}}\xspace}
\newcommand{\Observations}{\ensuremath{{\cal O}}\xspace}
\newcommand{\Actions}{\ensuremath{{\cal A}}\xspace}
\newcommand{\Transition}{\ensuremath{{\cal T}}\xspace}
\newcommand{\Observation}{\ensuremath{{\cal Z}}\xspace}
\newcommand{\Reward}{\ensuremath{{\cal R}}\xspace}
\newcommand{\Discount}{\gamma}

\newcommand{\History}[1]{\ensuremath{{h_{#1}}}\xspace}

\newcommand{\nrm}[2]{\ensuremath{{\parallel}{#2}{\parallel_{#1}}}}

\newcommand{\floor}[1]{\ensuremath{\lfloor{#1}\rfloor}}

\newcommand{\NormWeight}{\ensuremath{\hat{w}}\xspace}
\newcommand{\Weight}{\ensuremath{w}\xspace}
\newcommand{\WeightsSum}{\ensuremath{{\sum_{s_j \in \States} \Weight_j}}\xspace}
\newcommand{\tildeWeightsSum}{\ensuremath{{\sum_{s_j \in \States} \tilde{\Weight_j}}}\xspace}
\usepackage{algpseudocode}
\usepackage{svg}
\usepackage{xurl}
\title{
Anytime  Incremental $\rho$POMDP Planning in Continuous Spaces
}

\author{
Ron Benchetrit$^1$
\and
Idan Lev-Yehudi$^2$
\and
Andrey Zhitnikov$^3$
\and
Vadim Indelman$^{3,4}$\\
\affiliations
$^1$Department of Computer Science\\
$^2$Technion Autonomous Systems Program (TASP)\\
$^3$Department of Aerospace Engineering\\
$^4$Department of Data and Decision Sciences\\
Technion - Israel Institute of Technology, Haifa 32000, Israel\\
\emails
ronbenc@campus.technion.ac.il,
vadim.indelman@technion.ac.il
}

\begin{document}

\maketitle

\begin{abstract}

	Partially Observable Markov Decision Processes (POMDPs) provide a robust framework for decision-making under uncertainty in applications such as autonomous driving and robotic exploration. 
	Their extension, $\rho$POMDPs, introduces belief-dependent rewards, enabling explicit reasoning about uncertainty. 
	Existing online $\rho$POMDP solvers for continuous spaces rely on fixed belief representations, limiting adaptability and refinement - critical for tasks such as information-gathering. 
	We present $\rho$POMCPOW, an anytime solver that dynamically refines belief representations, with formal guarantees of improvement over time. 	
	To mitigate the high computational cost of updating belief-dependent rewards, we propose a novel incremental computation approach. We demonstrate its effectiveness for common entropy estimators, reducing computational cost by orders of magnitude.  
	Experimental results show that $\rho$POMCPOW outperforms state-of-the-art solvers in both efficiency and solution quality.
	
\end{abstract}

\section{Introduction}
	Autonomous agents must make informed decisions in the face of uncertainty, stemming from both the environment's dynamics and the agent's perception. 
	Sources of uncertainty include sensor noise, modeling approximations, and stochastic changes in the environment over time. 
	A common framework for addressing these challenges is the Partially Observable Markov Decision Process (POMDP).

	In POMDPs, decision-making relies on the history of past actions and observations, but storing this history over long trajectories is impractical. 
	Instead, beliefs—probability distributions over unobserved states—serve as sufficient statistics, encoding all necessary information for optimal decision-making \cite{Thrun05book}.
	A solution to a POMDP is a policy that maps each belief to an action that maximizes the expected sum of future rewards. 
	However, finding exact solutions is computationally infeasible except for trivial cases \cite{Papadimitriou87math}, prompting the development of approximate algorithms.

	Tree search algorithms are a prominent method for approximating solutions to POMDPs and are the focus of this work. 
	Instead of evaluating all possible belief states—an infeasible task due to the immense size of the belief space—these algorithms concentrate on the subset of beliefs that can be reached from the initial belief through a sequence of actions and observations.
	In the online paradigm, a planner builds a search tree at each time-step to determine an approximately optimal action. Anytime algorithms are particularly valuable in this setting, as they can provide progressively better solutions as computation time permits.

	A POMDP with state-dependent rewards addresses uncertainty only indirectly, limiting its suitability for tasks like uncertainty reduction and information gathering.
	In contrast, belief-dependent rewards—defined over the state distribution—offer a more intuitive and effective framework for these tasks.
	For example, in active localization \cite{Burgard97ijcai}, the goal is to minimize uncertainty about the agent's state rather than reaching a specific destination.
	Belief-dependent rewards have also been applied to problems with sparse rewards, aiding decision-making through reward-shaping techniques \cite{Flaspohler19ral,Fischer20icml}
	 or as heuristics for guiding tree search \cite{do2023information}.

	When the reward depends on the belief, the problem extends to a $\rho$POMDP \cite{Araya10nips}, also known as Belief Space Planning (BSP) \cite{Platt10rss,VanDenBerg12ijrr,Indelman15ijrr}.
	Belief-dependent rewards are often derived from information-theoretic measures such as entropy or information gain.
	However, calculating these rewards for general continuous distributions is infeasible and relies on costly sample-based methods, including kernel density estimation and particle filtering \cite{Boers10fusion}.
	These methods are computationally expensive, with costs scaling quadratically with the number of samples, and they provide only asymptotic guarantees—requiring a large number of samples to achieve sufficient accuracy.

	In this paper, we address the challenge of planning in continuous-spaces $\rho$POMDPs. We introduce $\rho$POMCPOW, an anytime online solver for $\rho$POMDPs that incrementally refines belief representations with formal guarantees of improvement over time and efficiently performs incremental computation of belief-dependent rewards.  
	Before detailing our contributions, we review the most relevant prior work.

\section{Related Work}
	We will first review the most relevant online POMDP solvers, followed by those specifically designed for $\rho$POMDPs.
	\subsection{State of the art online POMDP solvers}

\begin{figure}[tb]
			\centering
			\includesvg[width=0.4\textwidth]{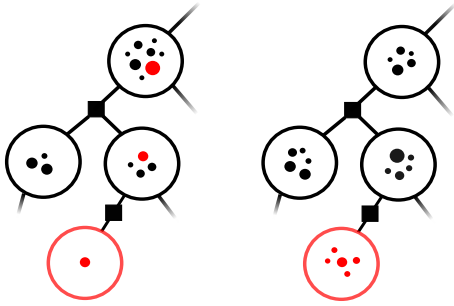}
			\caption{Illustration of belief tree construction by a state simulator (left) and a belief simulator (right). New particles and new nodes are marked in red. The state simulator updates beliefs by adding new particles along the trajectory, while the belief simulator maintains fixed beliefs once created.}
			\label{fig:trees}
		\end{figure}

		Online POMDP solvers for large or infinite state spaces typically approximate the belief using a set of state samples, commonly referred to as particles. 
		This particle-based representation is flexible and well-suited for capturing complex, multimodal beliefs \cite{Thrun05book}.
		For discussion purposes, these solvers can be broadly categorized into two groups: state simulators and belief simulators.
		See Figure \ref{fig:trees} for a simple illustration.

		\textbf{State Simulators:}
		State simulators focus on simulating state trajectories directly, incrementally updating visited beliefs with new particles at each visitation.
		Examples of state simulators include:
  
		POMCP \cite{Silver10nips}, which extends the UCT algorithm \cite{Kocsis06ecml} to the POMDP framework.  
		DESPOT \cite{Somani13nips} and its successors \cite{Ye17jair,Garg19rss}, which use heuristics to guide the search process.  
		POMCPOW \cite{Sunberg18icaps}, which extends POMCP to continuous action and observation spaces by incorporating progressive widening.  
		LABECOP \cite{Hoerger21icra}, an algorithm for continuous observation spaces, extracts the belief from scratch for each sampled observation sequence.

		A common trait of these algorithms is that each time a belief node is visited, the belief is updated with additional particles. 
		Intuitively, this approach improves the belief representation in frequently visited nodes, aligning with the exploration-exploitation trade-off.

		\textbf{Belief Simulators:}  
		Belief simulators, on the other hand, treat POMDP belief states as nodes in an equivalent Belief-MDP. Examples include:  

		PFT-DPW \cite{Sunberg18icaps}, which represents each belief node with a fixed number of particles. This makes the approach simple to implement and particularly effective for belief-dependent rewards, as rewards are computed once upon node creation.  
		AdaOPS \cite{Wu21nips}, which dynamically adapts the number of particles per belief node and aggregates similar beliefs, achieving competitive results compared to other state-of-the-art solvers.

		A key limitation of belief simulators is their fixed belief representation, which does not improve over time. This inefficiency leads to unpromising regions of the search space receiving the same computational effort as promising ones. 
		Moreover, these algorithms are less flexible when planning times vary.  Given ample time, belief simulators can construct dense trees, but belief representations at individual nodes may remain suboptimal. 
		Under time constraints, however, they often produce shallow, sparse trees, as significant computational effort is spent maintaining fixed belief representations rather than effectively exploring the search space.
				
\subsection{$\rho$POMDP Solvers}

	Several algorithms have been proposed to address the challenges of online planning in $\rho$POMDPs.

	PFT-DPW \cite{Sunberg18icaps} was introduced to accommodate belief-dependent rewards in POMDPs, though it was not demonstrated for this application. 
	Building on PFT-DPW, IPFT \cite{Fischer20icml} introduced the concept of reward shaping using information-theoretic rewards. It reinvigorates particles at each traversal of posterior nodes and estimates information-theoretic rewards by a kernel density estimator (KDE).
	
	AI-FSSS \cite{Barenboim22ijcai} reduces the computational cost of information-theoretic rewards by aggregating observations, providing bounds on the expected reward and value function to guide the search. Despite this improvement, its approach remains constrained by a fixed observation branching factor and a fixed number of particles per belief node. 
	\cite{Zhitnikov24ijrr} introduce an adaptive multilevel simplification paradigm for $\rho$POMDPs, which accelerates planning by computing rewards from a smaller subset of particles while bounding the introduced error. While their current implementation builds upon PFT-DPW, future extensions could complement our approach.
	
	All the above algorithms belong to the belief-simulators family and share the limitation of fixed belief representations.  

	An exception, closely related to our work, is $\rho$POMCP \cite{Thomas21arxiv}, which extends POMCP to handle belief-dependent rewards by propagating a fixed set of particles from the root instead of simulating a single particle per iteration. Their approach includes variants such as Last-Value-Update (LVU), which use the most recent reward estimates to reduce bias, unlike POMCP’s running average.

	However, $\rho$POMCP is limited to discrete spaces and recomputes belief-dependent rewards from scratch whenever a belief node is updated. This is costly in general and especially in continuous spaces, where the number of particles in the belief can grow indefinitely. These limitations highlight the need for efficient incremental updates to avoid full recomputation—an issue directly addressed by our approach.

\section{Contributions}

	To address the limitations of current state-of-the-art $\rho$POMDP solvers, we introduce $\rho$POMCPOW, an anytime solver for $\rho$POMDPs in continuous spaces. Our contributions are as follows:

\begin{itemize}
	\item \textbf{Algorithm Design:}  
	We introduce $\rho$POMCPOW, a novel anytime solver for $\rho$POMDPs in continuous spaces. $\rho$POMCPOW uses state simulations to explore the belief space, focusing computational resources on promising regions. 
	Its design improves belief representations over time, enabling more accurate and efficient computation of belief-dependent rewards.
	
	\item \textbf{Theoretical Foundations:}  
	We provide a general theoretical result establishing deterministic lower bounds on node visitation counts in online tree search algorithms. This result is applied in $\rho$POMCPOW to guarantee that belief representations improve as a function of time, ensuring progressively refined accuracy with continued planning.
			
	\item \textbf{Algorithmic Innovations:}  
	We introduce a novel incremental computation framework for belief-dependent rewards, integrated directly into $\rho$POMCPOW. Specifically, we demonstrate how to incrementally compute Shannon entropy and an entropy estimator proposed by \cite{Boers10fusion}, significantly reducing computational overhead compared to traditional full recomputation approaches.  
	
	Additionally, we present an incremental update mechanism for the Last-Value-Update (LVU) framework, further reducing computational costs by updating value estimates without full recomputation.

	\item \textbf{Empirical Validation:}  
	We conduct extensive experiments to demonstrate the advantages of $\rho$POMCPOW. The results show superior performance compared to state-of-the-art solvers in terms of computational efficiency, solution quality, and scalability.
\end{itemize}

\section{Background}
This section reviews mathematical formulations and notations used in this work.
\subsection{POMDPs}
A partially observable Markov decision process (POMDP) generalizes Markov decision processes (MDPs) to environments where the agent lacks full observability. Formally, a POMDP is defined by the tuple $\langle \States, \Actions, \Observations, \Transition, \Observation, \Reward, \Discount \rangle$, where:
\States, \Actions, and \Observations are the state, action, and observation spaces; $\Transition(s' | s, a)$ is the state transition model; $\Observation(o | a, s')$ is the observation model; $\Reward(s, a, s')$ is the state-dependent reward function; and $\Discount \in (0, 1]$ is the discount factor.

Since the agent cannot directly observe the state, it maintains a belief $b_t$, a probability distribution over states, updated using the history of past actions and observations. We denote the history at time $t$ as $\History{t} = (b_0, a_0, o_1, \dots, a_{t-1}, o_t)$, and the belief at time $t$ as $b_t(s) = \prob{s | \History{t}}$. The belief is a sufficient statistic for optimal decision-making in POMDPs \cite{Thrun05book}.

The agent’s objective is to find a policy $\pi$ that maps beliefs to actions and maximizes the expected sum of discounted rewards: $\pi^* = \argmax_\pi \mathbb{E}\left[ \sum_{t=0}^\infty \Discount^t \Reward(s_t, \pi(b_t), s_{t+1}) |b_0\right]$.

\subsection{$\rho$POMDPs}  
In this work, we focus on an extension of POMDPs, often referred to as $\rho$POMDPs, where the reward function depends on the belief state. We replace the state-dependent reward $\Reward$ with the belief-dependent reward $\rho$, structured as $\rho(b, a, b')$
The value function under policy $\pi$ is given by:  
\begin{align}
    V^\pi(b_0) &= \mathbb{E} \left[ \sum_{t=0}^\infty \Discount^t \rho(b_t, \pi(b_t), b_{t+1})\right],
\end{align}
where the expectation is over future beliefs. The corresponding action-value function satisfies:  
\begin{align}
    Q^\pi(b, a) &= \mathbb{E}_{b'} \left[ \rho(b, a, b') + \Discount V^\pi(b') \right].
\end{align}
For notational convenience, we may use \( hao \) to implicitly encode the relevant data of the belief \( b' \) resulting from history \( h \), action \( a \), and observation \( o \), enabling shorthand expressions such as \( \rho(hao) \) and \( V^\pi(hao) \).



\subsection{Online Tree Search}
Solving infinite-horizon POMDPs is computationally intractable. Online tree search approximates the optimal policy by constructing a search tree in real-time, starting from the current belief and exploring future actions and observations up to a predefined depth. However, constructing a full tree is infeasible due to the exponential growth of the search space—commonly known as the curse of history.

Monte Carlo Tree Search (MCTS) addresses this by iteratively performing four steps: selection, where the tree is traversed using a strategy like UCB \cite{Kocsis06ecml} to balance exploration and exploitation; expansion, which adds new child nodes to the tree; simulation, where trajectories are simulated from the expanded nodes to estimate values; and backpropagation, which updates statistics along the path from the leaf to the root. This approach builds an asymmetric tree focused on promising regions of the search space and provides anytime solutions.

POMCP \cite{Silver10nips} extends UCT \cite{Kocsis06ecml} to POMDPs by representing beliefs as particle sets and propagating state particles through the tree. This enables scalable planning for large discrete POMDPs but struggles in continuous spaces, where each simulation tends to create a new branch, resulting in sparse and shallow trees.

POMCPOW \cite{Sunberg18icaps} extends POMCP to continuous action and observation spaces using progressive widening, which limits the creation of new branches and allows existing branches to accumulate particles. It also employs a weighted particle filter to mitigate particle degeneracy, ensuring a robust belief representation.

\section{$\rho${POMCPOW}}

We introduce $\rho$POMCPOW, an online tree search algorithm for solving $\rho$POMDPs with continuous state, action, and observation spaces. $\rho$POMCPOW extends POMCPOW by incorporating belief-dependent rewards and modifying the backpropagation step to adopt the Last-Value-Update (LVU) framework \cite{Thomas21arxiv}, ensuring that only the latest reward estimates are used in value updates.

\subsection{Algorithm Overview}
Similar to POMCPOW, $\rho$POMCPOW iteratively constructs a search tree by simulating state trajectories through alternating layers of action and observation nodes. Each iteration begins by sampling a state from the initial belief and traversing the tree using predefined selection strategies, such as progressive widening for continuous spaces. State particles are updated along the trajectory by simulating actions, weighting by the observation model, and resampling to maintain a representative particle distribution. The process continues until either a depth limit is reached or a leaf node is encountered, at which point a rollout is performed to estimate the node’s value.

Unlike POMCPOW, $\rho$POMCPOW supports belief-dependent rewards, requiring modifications to the backpropagation step. Instead of using the classical Monte Carlo running average, which aggregates cumulative state-dependent rewards, $\rho$POMCPOW updates node values based on the most recent estimates from child nodes. While the LVU framework was introduced in \cite{Thomas21arxiv}, $\rho$POMCPOW further differs by implementing an incremental update mechanism for both value and action-value estimators. This mechanism efficiently adjusts estimates without recalculating them from scratch, significantly reducing computational overhead, particularly in continuous spaces. For full derivations of this novel incremental update mechanism, see Appendix~\ref{appendix:incrementalLVU}.

A key conceptual difference is that $\rho$POMCPOW simulation returns explicit value and action-value estimates rather than cumulative rewards along a trajectory. This adjustment is necessary for handling belief-dependent rewards and aligns with the LVU framework’s emphasis on using the most recent reward estimates. To reflect these changes, $\rho$POMCPOW restructures the simulation process into two procedures: \texttt{SimulateV} for propagating value estimates and \texttt{SimulateQ} for updating action-value estimates. 

Algorithm \ref{alg:rhoPOMCPOW} details the framework.

\begin{algorithm}[tb] 
	\caption{LVU $\rho$POMCPOW}
	\label{alg:rhoPOMCPOW}
	\begin{algorithmic}[1]
		\Procedure{SimulateV}{$s, h, d$}
		\If{$d = 0$}
			\State \Return 0
		\EndIf
		\State $a \gets \text{ActionSelection}(\dots)$
        \State $Q^{prev}(ha) \gets Q(ha)$
		\State $Q(ha) \gets \text{SimulateQ}(s, ha, d)$
		\State $N(h) \gets N(h) + 1$
		\State $V(h) \gets V(h) + \frac{1}{N(h)}[N(ha)Q(ha) -$
        \Statex \hspace{12mm} $(N(ha)-1)Q^{prev}(ha) - V(h)]$
		\State \Return $V(h)$
		\EndProcedure 
		\\
		\Procedure{SimulateQ}{$s', ha, d$} 
		\State Sample $s' \sim \Transition(\cdot|s, a)$
		\State $o \gets \text{ObservationSelection}(\dots)$
		\State Append $s'$ to belief $B(hao)$
		\State Append $\Observation(o| a, s')$ to weights $W(hao)$
		\State $\rho^{prev}(hao) \gets \rho(hao)$, $V^{prev}(hao) \gets V(hao)$
		\State $\rho(hao) \gets \text{UpdateReward}(\dots)$ \label{line:updatereward}
		\If{$o \notin C(ha)$} \Comment{new node}
			\State $C(ha) \gets C(ha) \cup \{o\}$
			\State $V(hao) \gets \text{ROLLOUT}(s', hao, d-1)$
		\Else
			\State Select $s' \in B(hao)$ based on weights $W(hao)$
			\State $V(hao) \gets \text{SimulateV}(s', hao, d-1)$
		\EndIf
		\State $N(ha) \gets N(ha) + 1$
        \State $Q(ha) \gets Q(ha) + \frac{N(hao)}{N(ha)}\left[\rho(hao)+\gamma V(hao)\right] -$
        \Statex \hspace{2mm} $\frac{N(hao)-1}{N(ha)}\left[\rho^{prev}(hao)+\gamma V^{prev}(hao)\right]-\frac{1}{N(ha)}Q(ha)$
        \State \Return $Q(ha)$
		\EndProcedure
	\end{algorithmic}
\end{algorithm}

The \texttt{ActionSelection} and \texttt{ObservationSelection} functions are abstracted for flexibility and can incorporate strategies such as the ones presented in \cite{Sunberg18icaps}. 

To enhance the accuracy of initial belief-dependent rewards and enable rollouts with belief-dependent rewards, a potential extension involves propagating a "bag of particles"—a fixed set of particles initialized at the root node and carried through each tree traversal, as suggested in \cite{Thomas21arxiv}. While promising, this extension lies beyond the scope of this work and is left for future exploration.




\subsection{Challenges and Discussion}

The $\rho$POMCPOW algorithm introduces a novel approach for solving $\rho$POMDPs, but two critical aspects require deeper exploration to fully harness its capabilities.

First, belief representation within the search tree is vital for tasks such as information gathering, where accurately modeling uncertainty is essential.
While the algorithm accumulates particles in belief nodes based on visitation counts, this does not guarantee adequate representation across the belief tree. 
Some belief nodes may remain underrepresented, limiting the algorithm’s effectiveness. In the next section, we formally analyze this behavior and prove that, under action and observation selection strategies satisfying a specific property, the belief representation of each node improves over time, leading to a more accurate depiction of the belief space.

Second, since $\rho$POMCPOW updates each visited belief along the simulated trajectory, belief-dependent rewards must also be updated, posing a significant computational challenge. These rewards are typically non-linear functions of the belief, making efficient updates non-trivial.
Recomputing rewards from scratch for every new particle is expensive, particularly in continuous state spaces, where the number of particles grows unbounded. For instance, both KDE and \cite{Boers10fusion} entropy estimators scale quadratically with the number of particles. In a subsequent section, we introduce an incremental belief-dependent reward computation, demonstrating how it enables efficient updates for various reward functions.


\section{Visitation Count and Belief Refinement}

Online tree search algorithms must balance broad exploration of the search space with refining estimates of existing nodes, a challenge known as the exploration-exploitation trade-off.

Algorithms such as UCT and POMCP address this trade-off by adopting the principle of optimism in the face of uncertainty, where actions are assumed to be promising until sufficient evidence suggests otherwise.

This challenge is amplified in continuous spaces, where the search tree can grow indefinitely, making it difficult to ensure adequate visitation across all nodes. 
In fact, it has been shown that in POMCP-DPW \cite{Sunberg18icaps}, when operating in continuous spaces, posterior nodes in the belief tree are visited only once, severely limiting their representation and hindering uncertainty estimation.

POMCPOW mitigates this issue through Progressive Widening, which controls the expansion rate of new child nodes. However, its effect on node visitation has not been formally analyzed, leaving open questions about how well belief representations improve over time.

To address this, we introduce the concept of a consistent selection strategy to ensure 
sufficient node visitations for each node and derive a deterministic lower bound on 
visitation counts. While broadly applicable to tree search, we demonstrate 
its use in $\rho$POMCPOW, guaranteeing improved belief representations over time.


\subsection{Consistent Selection Strategies}

Let $N(v; t)$ denote the visitation count of node $v$ at the $t^{th}$ iteration of the algorithm. We define a consistent selection strategy as follows:
	
\begin{definition}[Consistent Selection Strategy]
\label{definition:consistentselection}
A selection strategy is consistent if there exist non-decreasing functions \( f \) and \( F \), where \( \lim_{n \to \infty} F(n) = \infty \), such that for any node \( v \) with \( N(v; t) \geq f(i) \), the visitation count of its \( i^{th} \) child \( vi \) satisfies:
\begin{align}
N(vi; t) \geq F(N(v; t)).
\end{align}
\end{definition}
This definition ensures that once a parent node has been visited sufficiently often, its child nodes are also visited proportionally. The function \( F \) guarantees that visitation counts grow over time, enabling all parts of the tree to be explored adequately. 
In Example~\ref{example:augerconsistent}, we provide specific instances of consistent selection strategies, derived from the work of \cite{Auger13Sp}.

\subsection{Node Visitation Lower Bound}

Using consistent selection strategies, we establish a deterministic lower bound on the visitation count of each node in the belief tree.

\begin{theorem}[Node Visitation Lower Bound]
\label{theorem:nodevisitation}
Assume the action and observation selection strategies are consistent with functions \( f, F \) and \( g, G \), respectively. 
For a belief tree path \( h_{\tau} \), the visitation counts satisfy:
\begin{itemize}
    \item For \( h_{\tau} = a_{i_0}o_{j_1}a_{i_1}o_{j_2} \dots a_{i_{\tau-1}}o_{j_{\tau}} \), with \( t \geq k(i_0, j_1, \dots, i_{\tau-1}, j_{\tau}) \):
    \begin{align}
    N(h_{\tau}; t) \geq K_{\tau}(t) =  \underbrace{G \circ F \circ \dots \circ G}_{\tau \text{ times}}(t).
    \end{align}
\end{itemize}
Here, \( k(i_0, \dots, j_{\tau}) \) ensures sufficient initial visitation counts.
A more detailed version of this theorem, including the explicit closed-form expression for \( k \) and its complete proof, is provided in Appendix~\ref{appendix:selectionproof}. 
\end{theorem}

\begin{remark}
Since \( F \) and \( G \) are non-decreasing and \( \lim_{n \to \infty} F(n) = \lim_{n \to \infty} G(n) = \infty \), it follows that: 
\begin{align}
\lim_{t \to \infty} N(h_{\tau}; t) = \infty.
\end{align}
This guarantees that visitation counts grow indefinitely over time, ensuring sufficient exploration of nodes in the belief tree as the algorithm progresses.
\end{remark}

\begin{example}\label{example:augerconsistent}
	Algorithms~\ref{alg:augeractionselection} and \ref{alg:augerobservationselection} from \cite{Auger13Sp} establish a consistent selection strategy for action and observation nodes. 
	A detailed proof, the algorithm descriptions, and specific instances of the functions
	\(f\), \(F\), \(g\), and \(G\) are provided in Appendix~\ref{appendix:augerconsistent}.
	\end{example}

\subsection{Anytime Belief Refinement in $\rho$POMCPOW}

We now demonstrate how Theorem~\ref{theorem:nodevisitation} ensures that belief representations in $\rho$POMCPOW improve over time.

\begin{corollary}[Anytime Belief Refinement]
\label{corollary:beliefimprovement}
In $\rho$POMCPOW, under consistent action and observation selection strategies, Theorem~\ref{theorem:nodevisitation} guarantees that the visitation count of each node increases over time. Consequently, belief representations improve as planning progresses.
\end{corollary}

Thus, using selection strategies like those in Example~\ref{example:augerconsistent}, $\rho$POMCPOW overcomes the limitations of fixed-particles approaches by guaranteeing progressively refined belief representations, leading to more accurate modeling of the belief space.
While this property may or may not hold for other online POMDP solvers in continuous spaces, to the best of our knowledge, this is the first formal proof of such a property.

\section{Incremental Reward Computation}

As discussed in previous sections, updating belief-dependent rewards from scratch each time a belief is updated is prohibitively inefficient, particularly in continuous spaces where the number of particles grows indefinitely.

Most common belief-dependent rewards are information-theoretic and rely on entropy. We demonstrate incremental computation of Shannon entropy and the entropy estimator proposed by \cite{Boers10fusion}, referred to as the Boers entropy estimator, significantly reducing computational overhead compared to full recomputation.

These principles are general and extend to other belief-dependent rewards. For example, \cite{IKDE2020} proposes an incremental KDE update, which could be applied to KDE-based entropy estimators used in \cite{Fischer20icml}. 
Additionally, they may benefit other algorithms, such as $\rho$POMCP.

We represent the belief as a set of particles \( \{s_i, \NormWeight_i\}_{i=1}^N \), where \( s_i \) is a state sample and \( \NormWeight_i \) is the normalized weight of the particle. The normalized weight is computed as $\NormWeight_i = \frac{\Weight_i}{\sum_{j} \Weight_j}$,
where \( \Weight_i \) is the unnormalized weight of particle \( s_i \).


\subsection{Incremental Computation of Shannon Entropy}


While Shannon entropy is traditionally defined for discrete distributions, which are not the focus of this work, we present an incremental computation method tailored for particle-based belief representations to illustrate the broader feasibility of incremental computation for belief-dependent rewards.

The Shannon entropy of the particle belief is defined as:
\begin{align}
	\hat{H}(b) = -\sum_{s_i \in \States} \NormWeight_i \log \NormWeight_i.
\end{align}
Recomputing the entropy from scratch after introducing new particles incurs a computational cost proportional to the total number of particles, \( O(N) \). As the number of particles grows, this cost can become prohibitive. To address this, we use the following factorization:
\begin{align}
	\hat{H}(b) = -\frac{1}{\WeightsSum} \sum_{s_i \in \States} \Weight_i \log \Weight_i + \log(\WeightsSum),
\end{align}
When a new particle \( s_k \) is introduced, only \( \Weight_k \) changes, while other weights remain unchanged.\footnote{We assume identical state particles are merged and their weights added.}
By caching the previous entropy \( \hat{H}(b) \) and the sum of weights \( \WeightsSum \), the entropy can be incrementally updated in \( O(1) \) time, avoiding the need of recalculating from scratch. 
For detailed derivations of the incremental update formula, see Appendix~\ref{appendix:shannonentropy}.

\subsection{Incremental Computation of Boers Entropy Estimator}

While Shannon entropy provides a simple way to capture uncertainty, it is not suitable for continuous state spaces, which are the focus of this work. For a detailed discussion on its limitations in this context, see \cite{Boers10fusion}. 
The Boers entropy estimator is more suitable for this setting and is known to converge to the true entropy under mild assumptions.

However, the computational cost of the Boers entropy estimator scales quadratically with the number of particles, making it impractical to compute from scratch each time the belief is updated. 
To address this, we present a method to incrementally update the Boers entropy estimator, significantly reducing computational overhead.

The Boers entropy estimator is defined as:
\begin{align} \label{eq:log_decomp}
	\hat{H}(b') =& \log \left[ \sum\nolimits_{i=1}^{N}\Observation(o| a, s'_i)\NormWeight_{i} \right]
	- \sum\nolimits_{i=1}^{N}\NormWeight'_i \log \Observation(o| a, s'_i) \nonumber \\
    -& \sum\nolimits_{i=1}^{N}\NormWeight'_i \log \underbrace{\left[\sum\nolimits_{j=1}^{N}\Transition(s'_i|s_j, a)\NormWeight_j\right]}_{c_i},
\end{align}
where ( $'$ ) denotes quantities associated with the posterior belief, e.g., \( s'_i \) and \( \NormWeight'_i \) represent the state and normalized weight of the $i$th particle in the posterior belief, respectively. 
When the beliefs \( b \) and \( b' \) are updated with a new particle, the affected terms are denoted with a tilde, e.g., \( \tilde{w}_i \), \( \tilde{w}'_i \), and \( \tilde{c}_i \).

While all terms in Equation~\ref{eq:log_decomp} can be updated incrementally, we focus on the incremental update of the last, as it is the computational bottleneck of the Boers entropy estimator.
The updated \( \tilde{c}_i \) for \( i = 1, \dots, N \) is:
\begin{align} \label{eq:incremental_ci}
    \tilde{c}_i &= \sum\nolimits_{j=1}^{N}\Transition(s_i|s_j, a)\tilde{w}_j + \Transition(s_i|s_{N+1}, a)\tilde{w}_{N+1} \nonumber \\
    &= \frac{\sum\nolimits_{j=1}^{N}\Weight_j}{\sum\nolimits_{j=1}^{N+1}\Weight_j} c_i + \Transition(s_i|s_{N+1}, a)\tilde{w}_{N+1}.
\end{align}
Thus, by caching \( c_i \) and the sum of weights and reusing previously computed terms, Equation~\ref{eq:incremental_ci} can be updated in \( O(1) \) time for \( i = 1, \dots, N \).
For \( i = N+1 \), the term \( \tilde{c}_{N+1} \) must be computed from scratch, incurring \( O(N) \) cost. 
However, this computation is only required once for each new particle.
Overall, the Boers entropy estimator can be updated incrementally in \( O(N) \) significantly reducing computational cost.


\section{Experiments}


\urldef{\myurl}\url{https://github.com/ronbenc/Anytime-Incremental-rho-POMDP-Planning-in-Continuous-Spaces}

We implemented \(\rho\)POMCPOW in Julia via the POMDPs.jl framework~\cite{egorov2017pomdps}, with code on GitHub\footnote{\myurl}.

We evaluate it on two benchmark problems: the Continuous 2D Light-Dark problem and the Active Localization problem. We compare its performance against IPFT and PFT-DPW\footnote{We modified PFT-DPW to support belief-dependent rewards.}, two state-of-the-art \(\rho\)POMDP solvers, under varying planning time budgets. Performance is measured as the mean return with standard error over 1000 trials.

Next, we assess the impact of incremental reward computation by comparing planning time across iterations for \(\rho\)POMCPOW with and without incremental updates.

Detailed hardware specifications and solver hyperparameters used in the experiments are provided in Appendix~\ref{appendix:experimental}.

\subsection{Benchmark Problems}

Both problems share a common structure. The agent operates in a continuous 2D environment with an uncertain initial belief. Several beacons are scattered throughout the environment, and the agent receives noisy relative pose observations from the nearest beacon, where accuracy improves with proximity. The action space consists of movement in eight directions on the unit circle, along with a "stay" action that terminates the episode. State transitions are stochastic, and each step incurs a movement cost.

The key difference between the two problems lies in their objectives. In the Light-Dark problem, the agent aims to reach a goal region, while in Active Localization, the objective is to minimize uncertainty about its position.
For detailed parameters for both problems, see Appendix~\ref{appendix:problems}.
\subsubsection{Continuous 2D Light-Dark Problem}

In the 2D Light-Dark problem, the agent's task is to navigate toward a goal, starting with a highly uncertain initial belief.
The reward function is sparse, granting a large positive reward upon termination if the agent reaches the goal region and a large penalty otherwise. 
To reach the goal successfully, the agent must rely on beacons to improve localization.
To encourage this behavior, information gain, defined as $IG(b, b') = \hat{H}(b)-\hat{H}(b')$, is used as reward shaping, benefiting solvers that explicitly handle belief-dependent rewards.
We also evaluate POMCPOW without information gain to highlight $\rho$POMCPOW's ability to incorporate belief-dependent rewards.
Results are presented in Table~\ref{tab:cld2dresults}.
\begin{table}[tb]
    \centering
    \begin{tabular}{@{}lccc@{}}
    \toprule
    \multicolumn{4}{c}{Continuous 2D Light-Dark POMDP Scenario} \\ \midrule
    Algorithm               & 0.1 Seconds        & 0.2 Seconds          & 1.0 Seconds             \\ \midrule
    \textbf{$\boldsymbol{\rho}$POMCPOW}$^\dagger$ & \(\textbf{22.3} \pm 1.2\)   & \(\textbf{25.9} \pm 1.1\)     & \(\textbf{26.2} \pm 1.1\)        \\
    POMCPOW                 & \(17.2 \pm 1.4\)   & \(17.5 \pm 1.4\)     & \(18.5 \pm 0.9\)        \\
    IPFT$^\dagger$          & \(-2.3 \pm 1.8\)  & \(6.4 \pm 1.7\)      & \(17.2 \pm 1.2\)        \\
    PFT-DPW$^\dagger$       & \(5.3 \pm 1.6\)    & \(13.4 \pm 1.4\)     & \(20.5 \pm 1.0\)        \\ \bottomrule
    \end{tabular}
    \caption{Performance comparison for the Continuous 2D Light-Dark POMDP scenario. 
    Algorithms marked with $^\dagger$ use information gain as reward shaping.}
    \label{tab:cld2dresults}
\end{table}

Results show that \(\rho\)POMCPOW finds better solutions significantly faster.
The comparison with POMCPOW highlights $\rho$POMCPOW's ability to utilize belief-dependent rewards and shows that while these rewards introduce expensive computations, potentially slowing down simulations, they can significantly aid in finding better policies.

\subsubsection{Active Localization Problem}

\begin{figure}[tb]
    \centering
    \includegraphics[width=0.396\textwidth]{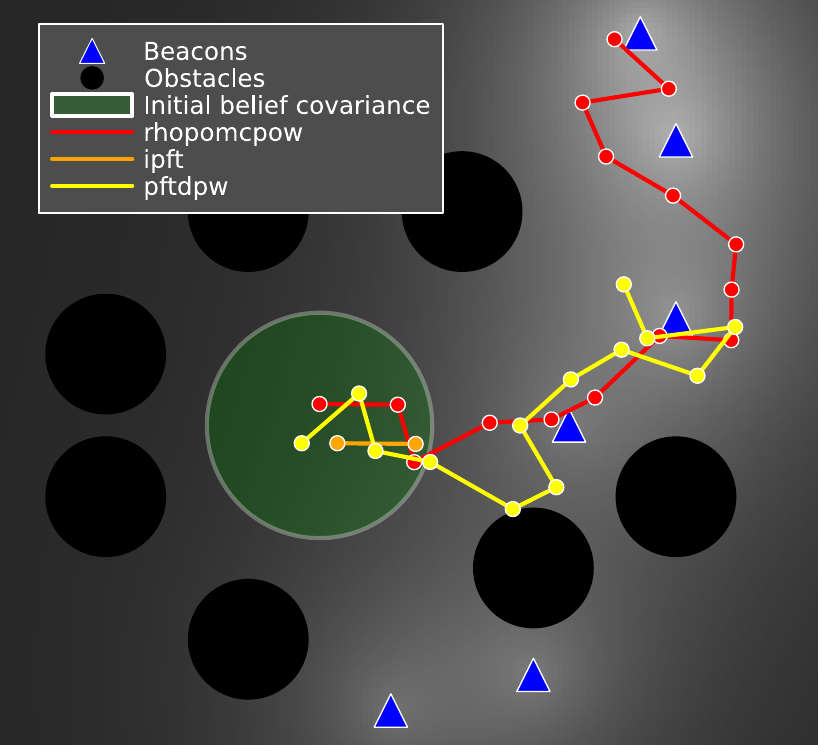}
	\caption{Simulated trajectories in the Active Localization problem}
    \label{fig:state_trajectories}
\end{figure}

The Active Localization problem tasks the agent with minimizing uncertainty about its position. As in the Light-Dark problem, the agent starts with a highly uncertain belief and must use beacon observations for localization. However, unlike Light-Dark, obstacles are scattered throughout the environment, incurring a penalty upon collision, and distant beacons provide more informative observations, making decision-making more challenging.
Figure \ref{fig:state_trajectories} shows simulated trajectories resulted from each solver in this problem.
Unlike Light-Dark, the reward is pure information gain, directly driving uncertainty reduction.
Results are presented in Table~\ref{tab:alresults}.

\begin{table}[tb]
    \centering
    \begin{tabular}{@{}lccc@{}}
    \toprule
    \multicolumn{4}{c}{Active Localization POMDP Scenario} \\ 
    \midrule
    Algorithm & 0.1 Seconds        & 0.2 Seconds        & 1.0 Seconds        \\ 
    \midrule
    \textbf{$\boldsymbol{\rho}$POMCPOW} & \(29.0 \pm 0.5\)     & \(\textbf{38.1} \pm 0.7\)     & \(\textbf{45.9} \pm 0.8\)     \\
    IPFT          & \(27.1 \pm 0.4\)     & \(27.7 \pm 0.4\)     & \(27.0 \pm 0.4\)     \\
    PFT-DPW       & \(\textbf{36.7} \pm 0.7\)     & \(37.6 \pm 0.7\)     & \(38.7 \pm 0.7\)     \\
    \bottomrule
    \end{tabular}
    \caption{Performance comparison for the Active Localization POMDP scenario.}
    \label{tab:alresults}
\end{table}
Initially, $\rho$POMCPOW is outperformed by PFT-DPW but surpasses it as planning progresses. We attribute this to poor initial belief nodes that refine over time.
IPFT lags behind; running the same problem without obstacles suggests they may be the cause (see Appendix~\ref{appendix:alnoobs}).

\subsection{Effect of Incremental Reward Computation}

To evaluate the advantage of incremental reward computation, we compare the planning time of \(\rho\)POMCPOW with and without incremental updates in the Continuous 2D Light-Dark problem. Using the same random seed and parameters for both variants ensures identical search tree expansion, isolating the impact of incremental updates on efficiency.
Figure~\ref{fig:incrementalreward} presents planning time as a function of iterations. The results reveal a significant performance gap, with full recomputation scaling at a higher order, demonstrating that incremental updates are crucial for $\rho$POMCPOW to be scalable.

\begin{figure}[tb]
    \centering
    \includegraphics[width=0.5\textwidth]{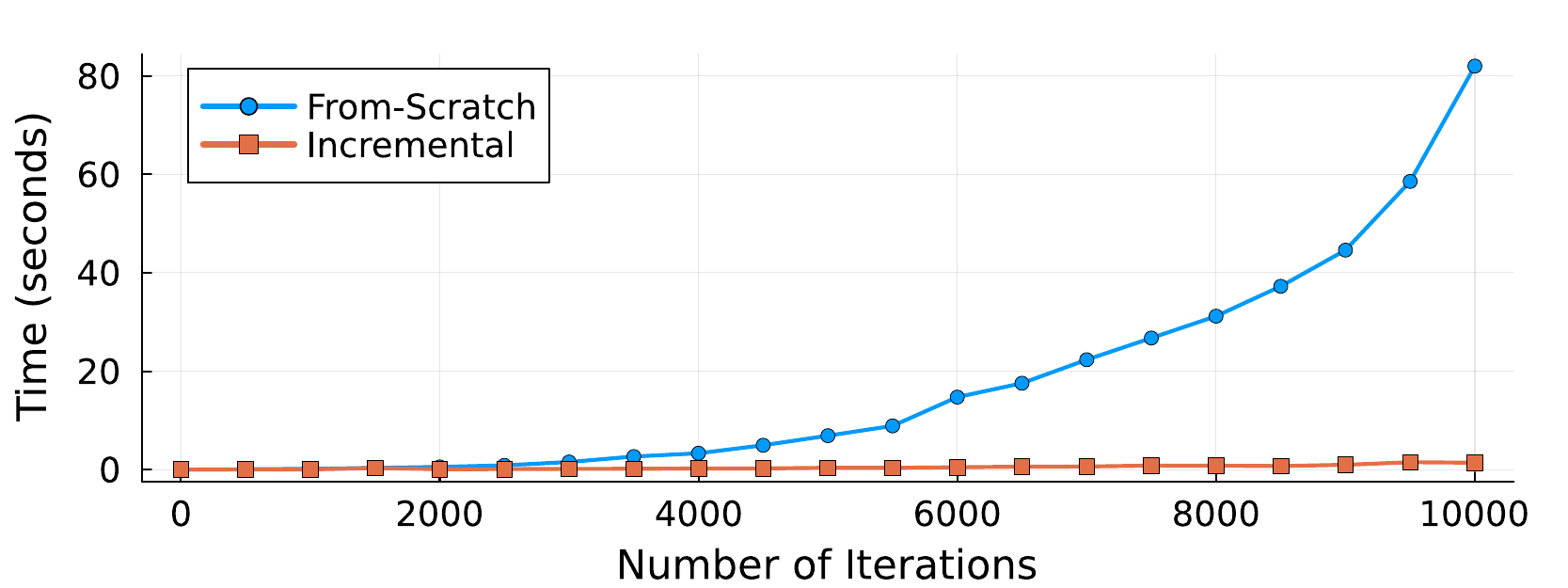}
	\caption{Planning time comparison for \(\rho\)POMCPOW with and without incremental reward computation as a function of iterations.}
    \label{fig:incrementalreward}
\end{figure}

A complexity analysis and a similar comparison with POMCPOW, assessing the cost of belief-dependent rewards, are presented in Appendix~\ref{appendix:beliefdependent}.

\section{Conclusions}
In this work, we introduced $\rho$POMCPOW, an online tree search algorithm for solving $\rho$POMDPs in continuous spaces.
Its belief representations are dynamic and guaranteed to improve over time, adressing a key limitation of state-of-the-art $\rho$POMDP solvers. 
Additionally, we introduced incremental reward computation and demonstrated its effectiveness on common entropy estimators, significantly reducing the computational overhead of belief-dependent rewards and enhancing scalability to continuous spaces. 
Experimental results highlight the necessity of incremental reward computation and validate $\rho$POMCPOW’s effectiveness in both solution quality and computational efficiency.


While the convergence of the algorithm remains an open question, 
\cite{Lim23jair} establish a link between particle count and performance guarantees, emphasizing the importance of improving belief representations over time. This is particularly crucial for belief-dependent rewards, where convergence guarantees are typically provided only in the limit. Theorem~\ref{theorem:nodevisitation} takes a key step in this direction by ensuring that beliefs refine over time, but a full convergence analysis is left for future work.

Despite efficient incremental updates, belief-dependent rewards remain the main computational bottleneck in $\rho$POMDP solvers (Appendix \ref{appendix:beliefdependent}). Addressing this challenge through approximation, parallelization, or reward function simplification is a key direction for future research.

\section*{Acknowledgements}
This research was supported by the Israel Science Foundation (ISF).

\bibliographystyle{named}
\bibliography{refs}

\newpage
\onecolumn
\appendix

\section{Incremental Last Value Update} \label{appendix:incrementalLVU}
\cite{Thomas21arxiv} introduced the concept of Last-Value Update (LVU) for $\rho$POMDPs, which focuses on updating the value function of belief nodes based on the most recent reward estimates. To the best of our knowledge, \cite{Thomas21arxiv} does not provide an incremental update mechanism for the value and action-value functions within the LVU framework. 

In this section, we present an incremental update mechanism for these functions, which is crucial for efficiently handling belief-dependent rewards in $\rho$POMCPOW.

\subsection{Incremental Update of the Value Function} 

\cite{Thomas21arxiv} introduced value estimates for belief nodes, initialized with rollout values when \( N(h) = 1 \).\footnote{In practice, POMCP and POMCPOW initialize visitation counts as \( N(h) = 0 \). 
To ensure consistency in state-dependent rewards, \( N(hao) \) should be replaced with \( N(hao) + 1 \).
} This leads to the following formula for the value function of a node \( h \):
\begin{align}
	\hat{V}(h) = \frac{1}{N(h)} \left[ Rollout(h) + \sum_{a \in Ch(h)} N(ha) \hat{Q}(ha) \right],
\end{align}
where \( Ch(h) \) represents the set of action child nodes of node \( h \).

Assume that at iteration \( t \), node \( h \) is visited and action \( a' \) is selected. The updated value function is given by:
\begin{align}
	\hat{V}(h; t) &= \frac{1}{N(h; t)} \left[ Rollout(h) + \sum_{a \in Ch(h; t)} N(ha; t) \hat{Q}(ha; t) \right] \\
	&= \frac{1}{N(h; t)} \left[ Rollout(h) + \sum_{a \in Ch(h; t) \setminus a'} N(ha; t) \hat{Q}(ha; t) + N(ha'; t) \hat{Q}(ha'; t) \right] \\
	&= \frac{1}{N(h; t)} \left[ Rollout(h) + \sum_{a \in Ch(h; t) \setminus a'} N(ha; t-1) \hat{Q}(ha; t-1) + N(ha'; t) \hat{Q}(ha'; t) \right] \\
	&= \frac{1}{N(h; t)} \Big[ Rollout(h) + \sum_{a \in Ch(h; t-1)} N(ha; t-1) \hat{Q}(ha; t-1) \nonumber \\
	&\hspace{15mm} - N(ha'; t-1) \hat{Q}(ha'; t-1) + N(ha'; t) \hat{Q}(ha'; t) \Big] \\
	&= \frac{N(h; t-1)}{N(h; t)} \hat{V}(h; t-1) - \frac{N(ha'; t-1)}{N(h; t)} \hat{Q}(ha'; t-1) + \frac{N(ha'; t)}{N(h; t)} Q(ha'; t).
\end{align}

Thus, the value function is incrementally updated as:
\begin{align}
	\hat{V}(h) \leftarrow \hat{V}(h) + \frac{1}{N(h)} \Big[ N(ha') \hat{Q}(ha') - (N(ha') - 1) \hat{Q}^{prev}(ha') - \hat{V}(h) \Big],
\end{align}
where \( \hat{Q}^{prev}(ha') \) is the previous value of the action-value function for action \( a' \).

This incremental approach allows the value function to be updated in \( O(1) \), compared to \( O(|Ch(h)|) \) for recomputing the entire sum of child nodes. This efficiency is particularly beneficial in large trees with numerous child nodes.

\subsection{Incremental Update of the Action-Value Function}
	Similarly, the action-value function for a node $ha$ is given by
	\begin{align}
		\hat{Q}(ha) \leftarrow \frac{1}{N(ha)}\sum_{o \in Ch(ha)}N(hao)\left[\hat{\rho} (hao) + \gamma \hat{V}(hao)\right]
	\end{align}
	Where $Ch(ha)$ is the set of observation children nodes of node $ha$.
	Assume at iteration $t$ node $ha$ is visited and that observation $o'$ is selected. The updated action-value function is given by:  
	\begin{align}
		\hat{Q}(ha; t) &= \frac{1}{N(ha; t)} \sum_{o \in Ch(ha; t)}N(hao; t)\left[\hat{\rho}(hao; t) + \gamma\hat{V}(hao; t)\right]
		\\
		&= \frac{1}{N(ha; t)} \sum_{o \in Ch(ha; t)\setminus o'}N(hao; t)\left[\hat{\rho}(hao; t) + \gamma \hat{V}(hao; t)\right] +\frac{N(hao'; t)}{N(ha; t)}\left[\hat{\rho}(hao'; t) + \gamma\hat{V}(hao'; t)\right]
		\\
		&= \frac{1}{N(ha; t)} \sum_{o \in Ch(ha; t-1)}N(hao; t-1)\left[\hat{\rho}(hao; t-1) + \gamma \hat{V}(hao; t-1) \right]
		\notag \\ &\qquad - \frac{N(hao'; t-1)}{N(ha; t)} \left[\hat{\rho}(hao'; t-1) + \gamma\hat{V}(hao'; t-1)\right]  +\frac{N(hao'; t)}{N(ha; t)}\left[\hat{\rho}(hao'; t) + \gamma\hat{V}(hao'; t)\right]
		\\
		&= \frac{N(ha; t-1)}{N(ha; t)}\hat{Q}(ha; t-1) - \frac{N(hao'; t-1)}{N(ha; t)} \left[\hat{\rho}(hao'; t-1) + \gamma\hat{V}(hao'; t-1)\right]  + \notag \\ &\qquad \frac{N(hao'; t)}{N(ha; t)}\left[\hat{\rho}(hao'; t) + \gamma\hat{V}(hao'; t)\right]
	\end{align}

	Thus, the action-value function is incrementally updated as:
	\begin{align}
		\hat{Q}(ha) \leftarrow \hat{Q}(ha) + \frac{1}{N(ha)} \Big[ N(hao') \left[ \hat{\rho}(hao') + \gamma \hat{V}(hao') \right] \nonumber \\
		- (N(hao') - 1) \left[ \hat{\rho}^{prev}(hao') + \gamma \hat{V}^{prev}(hao') \right] - \hat{Q}(ha) \Big].
	\end{align}

	Similarly, this incremental update allows the action-value function to be updated in \( O(1) \) time, compared to \( O(|Ch(ha)|) \) for recomputing the entire sum of child nodes. This efficiency is particularly beneficial when the number of sampled observations is large.

	\section{Theorem~\ref{theorem:nodevisitation}} \label{appendix:selectionproof}
	\subsection{Extended Theorem}
		Assume the action and observation selection strategies are consistent with functions \( f, F \) and \( g, G \), respectively. 
		For belief tree paths \( h_{\tau}^{-} \) and \( h_{\tau} \), the visitation counts satisfy:
		For a belief tree path \( h_{\tau} \), the visitation counts satisfy:
		\begin{itemize}
			\item For \( h_{\tau}^{-} = a_{i_0}o_{j_1}a_{i_1}o_{j_2} \dots a_{i_{\tau-1}} \), with \( t \geq k(i_0, j_1, \dots, i_{\tau-1}) \):
			\begin{align}
			N(h_{\tau}^{-}; t) \geq K_{\tau}^{-}(t) = F \circ \underbrace{G \circ F \circ \dots \circ G}_{\tau -1 \text{ times}}(t)
			\end{align}
			\item For \( h_{\tau} = a_{i_0}o_{j_1}a_{i_1}o_{j_2} \dots a_{i_{\tau-1}}o_{j_{\tau}} \), with \( t \geq k(i_0, j_1, \dots, i_{\tau-1}, j_{\tau}) \):
			\begin{align}
			N(h_{\tau}; t) \geq K_{\tau}(t) =  \underbrace{G \circ F \circ \dots \circ G}_{\tau \text{ times}}(t)
			\end{align}
		\end{itemize}
		\begin{align}
			k(i_0, \dots, j_{\tau}) = \max \big\{K_{0}^{-1}(f(i_0)), \dots, K_{\tau}^{-1}(g(j_{\tau}))\big\}
		\end{align}
	\subsection{Proof of Theorem~\ref{theorem:nodevisitation}}
	\begin{proof}
		We will prove this by induction.
		\begin{itemize}

			\item \textbf{Base Case:} 
			For $h_0 = b_0$, the statement is trivial with $K_0(t) = t$ and $k_0 = 0$.
	
			\item \textbf{Induction Hypothesis:} 
			Assume that for some $\tau \geq 0$, the following holds:
			\begin{align}
				n(h_{\tau}; t) \geq K_{\tau}(t)
			\end{align}
			for all $t \geq k_{\tau}(i_0, j_1, ..., i_{\tau-1}, j_{\tau})$.
	
			\item \textbf{Induction Step:} 
			We need to show that the hypothesis holds for $\tau + 1$.
	
			\begin{itemize}
				\item For $h_{\tau+1}^{-} = h_{\tau}a_{i_{\tau}}$:
					\begin{align}
						n(h_{\tau+1}^{-}; t) \geq F(n(h_{\tau}; t)) \geq F(K_{\tau}(t)) = K_{\tau+1}^{-}(t)
					\end{align}
					Where the first inequality holds when $n(h_{\tau}; t) \geq f(i_{\tau})$.
					Which in particular holds when $n(h_{\tau}; t) \geq K_{\tau}(t) \geq f(i_{\tau})$ which is true for all $t \geq k_{\tau}$ and $t \geq K_{\tau}^{-1}f(i_{\tau})$.
					The second inequality holds by the induction hypothesis and the monotonicity of $F$ for all $t \geq k_{\tau}$.
					Overall, both inequalities hold for all $t \geq \max \{k_{\tau}(i_0, j_1, ..., i_{\tau-1}, j_{\tau}), K_{\tau}^{-1}(f(i_{\tau}))\}= k_{\tau+1}^{-}(i_0, j_1, ..., i_{\tau})$.
	
				\item For $h_{\tau+1} = h_{\tau+1}^{-}o_{j_{\tau+1}}$:
					\begin{align}
						n(h_{\tau+1}; t) \geq G(n(h_{\tau+1}^{-}; t)) \geq G(K_{\tau+1}^{-}(t)) = K_{\tau+1}(t)
					\end{align}
					Where the first inequality holds when $n(h_{\tau+1}^{-}; t) \geq g(j_{\tau+1})$. 
					Which in particular holds when $n(h_{\tau+1}^{-}; t) \geq K_{\tau+1}^{-}(t) \geq g(j_{\tau+1})$ which is true for all $t \geq k_{\tau+1}^{-}(i_0, j_1, ..., i_{\tau})$ and $t \geq K_{\tau+1}^{-1}g(j_{\tau+1})$.
					The second inequality holds by the induction hypothesis and monotonicity of $G$ for all $t \geq k_{\tau+1}^{-}(i_0, j_1, ..., i_{\tau})$.
					Overall, both inequalities hold for all $t \geq \max \{k_{\tau+1}^{-}(i_0, j_1, ..., i_{\tau}), K_{\tau+1}^{{-}^{-1}}(g(j_{\tau+1}))\}= k_{\tau+1}(i_0, j_1, ..., i_{\tau}, j_{\tau+1})$.
			\end{itemize}
		\end{itemize}
	
		By induction, the hypothesis holds for all $\tau \geq 0$.
	\end{proof}

	\section{Example of consistent selection strategies} \label{appendix:augerconsistent}

	\begin{algorithm}[]
	\caption{Consistent Action Selection Procedure}
	\label{alg:augeractionselection}
	\begin{algorithmic}[1]
		\If{$\floor{N(h)^{\alpha_a}} > \floor{N(h)-1}^{\alpha_a}$}
			\State Sample a new action \( a \) and add a new child \( ha \)
		\Else
			\State Select the child \( ha \) that maximizes the score:
			\begin{align*}
				sc(ha) = \hat{Q}(ha; t) + \sqrt{N(h)^{e(d)}/{N(ha)}}.
			\end{align*}
		\EndIf
	\end{algorithmic}
	\end{algorithm}

	\begin{algorithm}[]
		\caption{Consistent Observation Selection Procedure}
		\label{alg:augerobservationselection}
		\begin{algorithmic}[1]
			\If{$\floor{N(ha)^{\alpha_o}} > \floor{N(ha)-1}^{\alpha_o}$}
				\State Sample \( o \sim \Observation(\cdot|s, a, s') \) and add a new child \( hao \)
			\Else
				\State Select the least visited child \( hao \) of \( ha \)
			\EndIf
		\end{algorithmic}
	\end{algorithm}
	Algorithms \ref{alg:augeractionselection} and \ref{alg:augerobservationselection} constitute a consistent selection strategies with the functions f, F, g and G respectively:
	\begin{align}
		f(i) &= i^{\frac{1}{\alpha_a(1-\alpha_a)}}, 
		& G(n) &= \frac{n}{\lfloor n \rfloor^{\alpha_o}} - 1\\
		g(i) &= \left\lceil (i+1)^{\frac{1}{\alpha_o}} \right\rceil,
		& F(n) &= \frac{1}{4}\,n^{\,e(d)\,(1-\alpha_a)}.
	  \end{align}

	The consistency of Algorithm \ref{alg:augeractionselection} is established by Lemma 3 in \cite{Auger13Sp}, while the consistency of Algorithm \ref{alg:augerobservationselection} follows from Corollary 2 in \cite{Auger13Sp}.

	\section{Incremental Computation of Shannon Entropy} \label{appendix:shannonentropy}
	We will demonstrate the incremental update of the Shannon entropy for a particle belief.
	Assume the belief is updated with a new particle \( s_k \). Denote updated quantities with \( \tilde{} \), e.g., \( \tilde{b} \), \( \tilde{\NormWeight}_i \), and \( \tilde{\Weight}_i \). The updated Shannon entropy is given by:

	\begin{align}
		\hat{H}(\tilde{b}) &= -\sum_{s_i \in \States} \tilde{\NormWeight}_i \log \tilde{\NormWeight}_i
		\\
		&= -\sum_{s_i \in \States} \frac{\tilde{\Weight}_i}{\tildeWeightsSum} \log \frac{\tilde{\Weight}_i}{\tildeWeightsSum}
		\\
		&= -\frac{1}{\tildeWeightsSum}\sum_{s_i \in \States} \tilde{\Weight}_ i \log \tilde{\Weight}_ i + \log(\tildeWeightsSum)
		\\
		&= -\frac{1}{\tildeWeightsSum}\sum_{s_ i \in \States} \Weight_ i \log \Weight_ i - \frac{\tilde{\Weight}_ k \log \tilde{\Weight}_ k - \Weight_ k \log \Weight_ k}{\tildeWeightsSum} +  \log(\tildeWeightsSum)
		\\
		&= -\frac{1}{\tildeWeightsSum}\sum_{s_ i \in \States} \Weight_ i \log \frac{\Weight_ i}{\WeightsSum} - \frac{\tilde{\Weight}_ k \log \tilde{\Weight}_ k - \Weight_ k \log \Weight_ k}{\tildeWeightsSum} +  \log(\tildeWeightsSum) - \log(\WeightsSum)
		\\
		&= \frac{\WeightsSum}{\tildeWeightsSum}\hat{H}(b) - \frac{\tilde{\Weight}_ k \log \tilde{\Weight}_ k - \Weight_ k \log \Weight_ k }{\tildeWeightsSum} -\log(\frac{\WeightsSum}{\tildeWeightsSum})
	\end{align}

	Thus, the Shannon entropy can be incrementally updated as:
	\begin{align}
		\hat{H}(b) \leftarrow \frac{\WeightsSum}{\tildeWeightsSum}\hat{H}(b) - \frac{\tilde{\Weight}_ k \log \tilde{\Weight}_ k - \Weight_ k \log \Weight_ k }{\tildeWeightsSum} -\log(\frac{\WeightsSum}{\tildeWeightsSum})
	\end{align}

	By caching the previous entropy \( \hat{H}(b) \) and the sum of weights \( \WeightsSum \), the Shannon entropy can be updated in \( O(1) \) time. This approach avoids the need to recompute the entropy from scratch, which would require \( O(N) \) time.
	
\section{Experimental Details} \label{appendix:experimental}
All experiments were conducted on an 11th Gen Intel\textsuperscript{\textregistered} Core\texttrademark\ i9-11900K CPU (3.5,GHz) with 64,GB of RAM.

To ensure a fair comparison with POMCPOW, \(\rho\)POMCPOW adopts the same action and observation selection strategies as in \cite{Sunberg18icaps}.
Building on the proof of Lemma~3 from \cite{Auger13Sp}, one can show that the action selection strategy, which relies on the classical UCB algorithm, is a consistent selection strategy. 
We believe that the same holds---albeit in a probabilistic sense---for the observation selection strategy, but we leave the formal proof for future work.

\subsection{Continuous 2D Light-Dark Problem}

Because this problem has a relatively small action space, we do not use 
progressive widening for actions, making the parameters \(\alpha_a\) and
\(K_a\) unnecessary. We tuned \(\rho\)POMCPOW and PFT-DPW by performing a 
grid search over the exploration bonus \(C\) and the observation widening 
parameters \(K_o\) and \(\alpha_o\), using a seed different from the main 
experiment. For PFT-DPW, we additionally searched over \(m\), the number 
of particles. Due to the algorithms' similarity, the parameters found for 
\(\rho\)POMCPOW served as a starting point for POMCPOW, while those 
identified for PFT-DPW served as a starting point for IPFT.

$\rho$POMCPOW, IPFT and PFT-DPW used information-gain as reward shaping such that the reward is structured as:
\begin{align}
	\rho(b, a, b') = \mathbb{E}_{s, s'}[\Reward_s(s, a, s')] +\lambda \cdot IG(b, b')
\end{align}
with $\Reward_s(s, a, s')$ being the standard reward function defined by the problem, $\lambda = 30.0$ and $IG(b, b')$ being the information gain between the belief states $b$ and $b'$.

The chosen parameters are reported in table \ref{tab:lightdarkparams}.

\begin{table}[h]
	\centering
	\begin{tabular}{@{}lcccc@{}}
	\multicolumn{4}{c}{} \\ \midrule
	Algorithm               & $c$       & $k_o$     & $\alpha_o$   &$m$      \\ \midrule
	\textbf{$\boldsymbol{\rho}$POMCPOW}		& \(120\)   & \(6\)     & \(1/30\)     &-   \\
	POMCPOW                 & \(100\)   & \(4\)     & \(1/30\)     &-  \\
	IPFT         			& \(100\)   & \(3\)     & \(1/40\)     & \(20\) \\
	PFT-DPW      			& \(80\)    & \(3\)     & \(1/40\)     &\(50\)     \\ \bottomrule
	\end{tabular}
	\caption{Parameters for the Continuous 2D Light-Dark POMDP Scenario}
	\label{tab:lightdarkparams}
\end{table}

$\rho$POMCPOW uses a larger $c$, $k_o$, and $\alpha_o$ than PFT-DPW and IPFT. A possible explanation is because $\rho$POMCPOW is a state simulator, and therefore runs faster simulations, which allows for more exploration.

\subsection{Active Localization Problem}
Since both problems share common structure, we used the same parameters for the Active Localization problem as for the Continuous 2D Light-Dark problem with the only exception that now $\rho$POMCPOW uses a belief node initialization of 10 particles instead of a single particle.

\section{Detailed Problems Description}  \label{appendix:problems}
Both problems share a common structure. 
The agent can move in 8 directions on the unit circle or choose a "stay" action, which terminates the problem. 
Each step incurs a cost of $-1$, and transitions are linear-gaussian and are defined by:
\begin{align}
	\Transition(\cdot|s, a) = \mathcal{N}(s + a, \Sigma_{\Transition}), \quad  \Sigma_{\Transition} = \begin{bmatrix}
		0.1 & 0 \\
		0 & 0.1
	\end{bmatrix}
\end{align}
The agent receives noisy relative pose observations from the nearest beacon, with accuracy improving with proximity. The observation model is defined as:
\begin{align}
	\Observation(\cdot|s, a, s') = \mathcal{N}(x_b - s', \Sigma_{\Observation}), \quad  \Sigma_{\Observation} = \frac{\sqrt{2}}{2} \nrm{2}{x_b - s'} \begin{bmatrix}
		1 & 0 \\
		0 & 1
	\end{bmatrix} + \Sigma_{x_b}
\end{align}
where $x_b$ is the position of the nearest beacon, and $\Sigma_{x_b}$ changes between problems. 
The initial belief is given by:
\begin{align}
	b_0 = \mathcal{N}(x_0, \Sigma_{0}), \quad  \Sigma_0 = \begin{bmatrix}
		2.5 & 0 \\
		0 & 2.5
	\end{bmatrix}
\end{align}
where $x_0$ changes between the problems.
Both problem use a discount factor, $\Discount=0.95$.
\subsection{Continuous 2D Light-Dark Problem} \label{appendix:cld2d}
In the Continuous 2D Light-Dark Problem the agent taks is to reach a goal area.
The agent receives a large reward of $+100$ upon termination if it is within a unit circle center around the goal or a large penalty of $-100$ otherwise.
In this problem $\Sigma_{x_b}=0.5 \times I$ for all beacons.
\subsection{Active Localization Problem} \label{appendix:al}
In the Active Localization Problem the agent's goal is to minimize uncertainty about its position in a continuous 2D environment.
The reward is structured as:
\begin{align}
	\rho(b, a, b') = \mathbb{E}_{s, s'}[\Reward_s(s, a, s')] +\lambda \cdot IG(b, b')
\end{align} 
with $\Reward_s(s, a, s')$ being the standard reward function defined by the problem, $\lambda = 30.0$ and $IG(b, b')$ being the information gain between the belief states $b$ and $b'$.

In this problem $\Sigma_{x_b}=\frac{0.5}{\nrm{2}{x_b}} \times I$, meaning that beacons that are further from the origin have a lower observation noise, encouraging the agent to explore the environment.
Obstacles are scattered in the environment, and the agent incurs a large penalty of -50  for collisions.

\section{Active Localization Without Obstacles} \label{appendix:alnoobs}
Table \ref{tab:alresults} shows the results for the Active Localization problem without obstacles.
IPFT slightly outperforms $\rho$POMCPOW and PFT-DPW.
\begin{table}[h]
    \centering
    \begin{tabular}{@{}lccc@{}}
    \toprule
    \multicolumn{4}{c}{Active Localization POMDP Scenario W/O Obstacles} \\ \midrule
    Algorithm       & 0.1 Seconds          & 0.2 Seconds          & 1.0 Seconds          \\ \midrule
    \textbf{$\boldsymbol{\rho}$POMCPOW}  & \(57.0 \pm 0.2\)     & \(57.5 \pm 0.2\)     & \(57.3 \pm 0.1\)          \\
    IPFT            & \(\textbf{57.8} \pm 0.2\)    & \(\textbf{58.4} \pm 0.2\)   & \(\textbf{59.1} \pm  0.2\)          \\
    PFT-DPW         & \(52.4 \pm 0.4\)   & \(53.8 \pm 0.4\)     & \(55.1 \pm 0.4\)          \\ \bottomrule
    \end{tabular}
    \caption{Performance comparison for the Active Localization POMDP scenario without obstacles. 
    Bold values indicate the best result in each column.}
    \label{tab:alresults}
\end{table}

\section{Effect of Belief-Dependent rewards} \label{appendix:beliefdependent}
\subsection{Complexity Analysis}
We analyze the complexity of $\rho$POMCPOW with and without belief-dependent rewards to assess the advantage of incremental updates. Additionally, we examine vanilla POMCPOW to evaluate the impact of belief-dependent rewards.
Let $T$ be the iteration budget, $D$ the tree depth, and $R(N)$ the complexity of computing the belief-dependent reward function where $N$ is the number of particles in the belief.

Ignoring the branching of action and observation nodes, as well as rollouts, the number of particles in each node along the simulated trajectory is $O(t)$, where $t$ is the current iteration. Since belief-dependent reward computation is the main bottleneck, the complexity of running $\rho$POMCPOW with belief-dependent rewards for $T$ iterations is:

\begin{align}
	O(D \cdot R(1) + D \cdot R(2) + \dots + D \cdot R(T)) = O\left(D \sum_{t=1}^{T} R(t)\right).
\end{align}

We now assume that computing the belief-dependent reward from scratch has complexity $O(N^2)$, while incremental updates have complexity $O(N)$, as is the case for the KDE and Boers entropy estimators. 

Thus, the complexity of $\rho$POMCPOW when computing belief-dependent rewards from scratch is:

\begin{align}
    O\left(D \sum_{i=1}^{T} i^2\right) = O\left(D \cdot \frac{T(T+1)(2T+1)}{6}\right) = O(D \cdot T^3).
\end{align}

Similarly, with incremental belief-dependent rewards, the complexity reduces to:

\begin{align}
    O\left(D \sum_{i=1}^{T} i\right) = O\left(D \cdot \frac{T(T+1)}{2}\right) = O(D \cdot T^2).
\end{align}

For POMCPOW, state-dependent reward computation is $O(1)$, while the main computational bottleneck is particle resampling at each node, which takes $O(\log N)$ time.

Thus, the complexity of running POMCPOW for $T$ iterations is:
\begin{align}
    O\left(D \sum_{i=1}^{T} \log i\right) = O(D \cdot T \log T).
\end{align}

Although this is a simplified analysis, it aligns with our empirical results, highlighting the need for efficient incremental updates for belief-dependent rewards. However, even with such updates, belief-dependent rewards remain the main computational bottleneck, limiting the scalability of $\rho$POMCPOW and other $\rho$POMDP solvers. A more detailed analysis is left for future work.

\subsection{Empirical Runtime Comparison with POMCPOW}
We ran both POMCPOW and $\rho$POMCPOW on the Continuous 2D Light-Dark problem.
Both algorithms share the same parameters and random seeds. While $\rho$POMCPOW incrementally computes belief-dependent rewards, it does not use them. These factors ensure that both algorithms construct the same search tree.
\begin{figure}[h]
    \centering
    \includegraphics[width=0.8\textwidth]{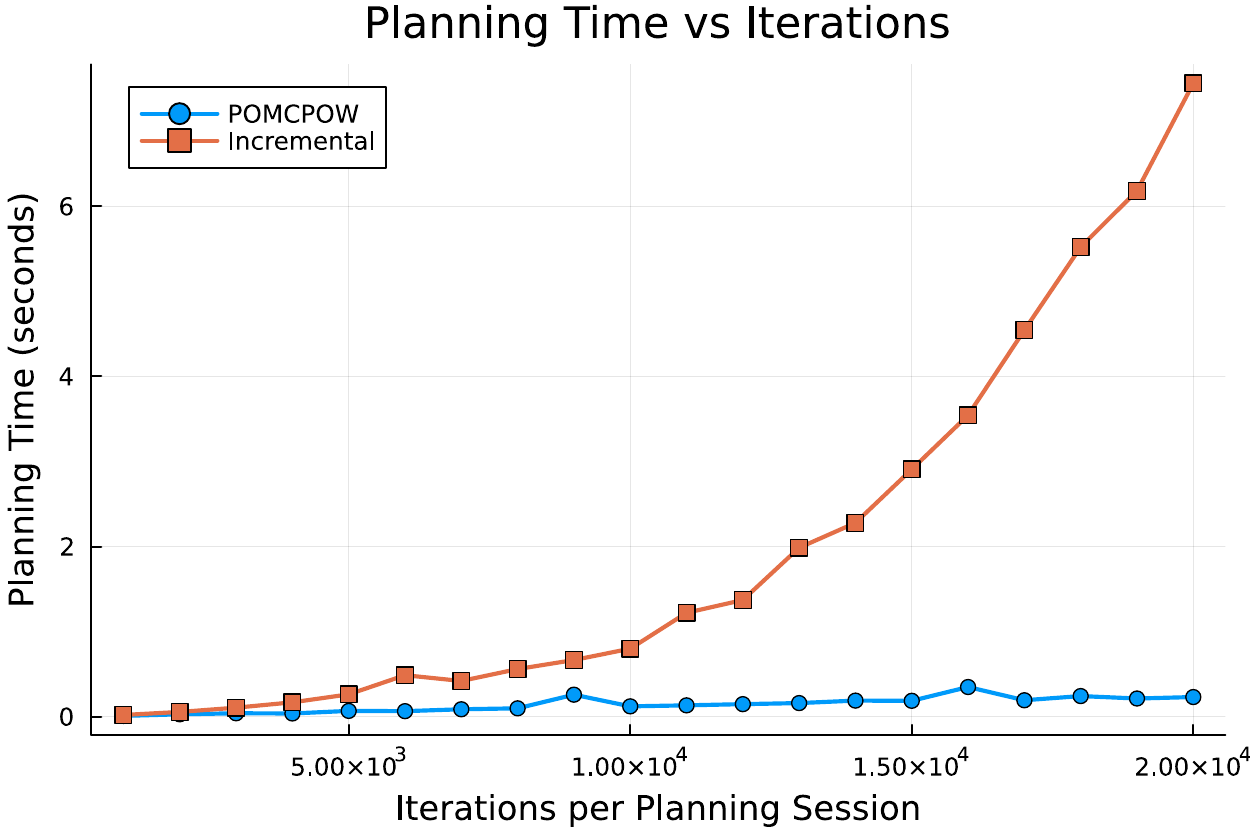}
	\caption{Planning time comparison for POMCPOW and \(\rho\)POMCPOW with incremental reward computation as a function of iterations.}
    \label{fig:incrementalreward}
\end{figure}

Results show that even with efficient incremental updates, belief-dependent rewards remain the primary computational bottleneck in $\rho$POMCPOW and other $\rho$POMDP solvers.
\end{document}